# Stop ordering machine learning algorithms by their explainability! A user-centered investigation of performance and explainability

Lukas-Valentin Herm [a,1], Kai Heinrich [b,2], Jonas Wanner [a,3], Christian Janiesch [c,*,4]

[a] *Julius-Maximilians-Universität Würzburg, Würzburg, Germany*
[b] *Otto-von-Guericke-Universität Magdeburg, Magdeburg, Germany*
[c] *TU Dortmund University, Otto-Hahn-Str. 14, 44227 Dortmund, Germany*

ABSTRACT

Machine learning algorithms enable advanced decision making in contemporary intelligent systems. Research indicates that there is a tradeoff between their model performance and explainability. Machine learning models with higher performance are often based on more complex algorithms and therefore lack explainability and vice versa. However, there is little to no empirical evidence of this tradeoff from an end user perspective. We aim to provide empirical evidence by conducting two user experiments. Using two distinct datasets, we first measure the tradeoff for five common classes of machine learning algorithms. Second, we address the problem of end user perceptions of explainable artificial intelligence augmentations aimed at increasing the understanding of the decision logic of high-performing complex models. Our results diverge from the widespread assumption of a tradeoff curve and indicate that the tradeoff between model performance and explainability is much less gradual in the end user's perception. This is a stark contrast to assumed inherent model interpretability. Further, we found the tradeoff to be situational for example due to data complexity. Results of our second experiment show that while explainable artificial intelligence augmentations can be used to increase explainability, the type of explanation plays an essential role in end user perception.

## 1. Introduction

Artificial intelligence (AI) technology enables human-like cognitive capacity for advanced decision making in contemporary intelligent systems (Dwivedi et al., 2021; Hradecky et al., 2022). Despite these advances, research shows that human perceptions can trigger behavioral responses that affect an organization's capability to leverage such systems when predictions are presented to end users (Berger et al., 2021; Chiu et al., 2021). As certain algorithms exhibit a higher degree of explainability through their inherent interpretability, end users may perceive them more benevolently than others (Shin, 2021; Wanner et al., 2022). In contrast, much of the current AI research focuses solely on the statistical performance measures of machine learning (ML) models (Collins et al., 2021; La Cava et al., 2019), and data competitions are dominated by deep neural network algorithms outperforming shallow ML algorithms (e.g., Hyndman, 2020; Rudin & Radin, 2019).

The processing of these deep neural network algorithms is based on complex calculation logic, which is practically untraceable. While the decision-making is documented in the learned model and could be traced, its inner complexity renders it unfeasible for humans to do so and interpret its decision-making process or even actual prediction results. It practically renders the model a black box as it does not provide any explanations for its predictions (Dwivedi et al., 2021; Loyola-Gonzalez, 2019). This results in a tradeoff between performance and explainability, which is not yet sufficiently understood from a user-centered perspective.

The performance of an algorithm can be measured by indicators such as accuracy, precision, recall, or the F-score. Yet, it remains unclear which ML algorithm's inherent interpretability is perceived as more explainable by end users. Even more so, explanations can be provided in several ways. Currently, it is unclear which types of explanation humans perceive more benevolently (Shin et al., 2020; von Eschenbach, 2021).

This is crucial as the perceived explainability of a prediction determines the effectiveness of an intelligent system: if human decision-makers can interpret the behavior of an underlying ML model, they are more willing to act based on it, especially in cases where the predictions do not conform to their own expectations (Berger et al., 2021; Ribeiro et al., 2016). Even more so, intelligent systems without sufficient explainability may even be inefficacious if end users disregard their advice (Shin et al., 2020; Wanner et al., 2021b).

In scholarly literature, several theoretical considerations on the tradeoff of performance and explainability exist (e.g., Arrieta et al., 2020; Dam et al., 2018; Gunning, 2019; James et al., 2013; Nanayakkara et al., 2018; Rudin, 2019; Yang & Bang, 2019). Yet, they are of theoretic nature, and an empirical investigation of model explainability is missing. We intend to close this gap with our first research question:

**RQ1:**.    *How do common classes of machine learning algorithms compare empirically in the tradeoff between their performance as measured by model accuracy and their explainability as perceived by end users?*

While we cannot increase the performance of individual ML models without modification of the actual analytics process (e.g., data preparation, hyperparameter tuning, etc.), any ML model's predictions can be augmented with external explanations. This is especially important for high-performing algorithms based on deep learning as they offer no inherent explainability to end users (Arrieta et al., 2020; Sharma et al., 2021). In response, several types of explainable AI (XAI) augmentations have been developed. Their visualizations can be grouped into several common types of explanations (Guidotti et al., 2018; Mohseni et al., 2021). There is little to no empirical evidence on actual end user perception when considering their role in intelligent system use (Hoffman et al., 2013; Hoffman et al., 2018). We formulate our second research question accordingly:

**RQ2:**.    *How do common types of explanations compare empirically in their explainability as perceived by end users?*

Our insights have a high potential to explain better AI adoption of different classes of ML algorithms, contributing to a better understanding of AI decision-making and the future of work. On the one hand, the results can help us to understand to what extent various classes of ML algorithms differ in their perceived explainability from an end user perspective. It allows us to draw conclusions about their future improvement and their suitability for a given situation in practice. On the other hand, the results can help us understand how much performance end users may be willing to forfeit in favor of explainability. Ultimately, Rudin (2019)'s call to avoid explaining black-box models in favor of using inherently interpretable white-box models could be better approached if the tradeoff was sufficiently understood from a social-technical, end user perspective (Arrieta et al., 2020; Herm et al., 2021).

The remainder of the paper is structured as follows: Section 2 introduces related work on ML algorithms and model explainability. In Section 3, we discuss the hypothesis development for our research questions, and in Section 4, we introduce our methodology as well as outline datasets and algorithms, implementation details, and survey design. In Section 5, we present the results of the empirical study before we discuss them in Section 6 and present implications. The paper closes with a brief summary.

## 2. Literature review

### 2.1. Machine learning algorithms

ML focuses on algorithms that improve their performance through experience. They are able to find non-linear relationships and patterns in datasets without being explicitly programmed to do so (Bishop, 2006; Russell & Norvig, 2021). The process of analytical modeling building to turn ML algorithms into concrete ML models for the use in intelligent systems is a four-step process comprising data input, feature extraction, model building, and model assessment (Goodfellow et al., 2016; Janiesch et al., 2021).

ML algorithms are commonly grouped into shallow and deep ML algorithms. Each ML algorithm has different strengths and weaknesses regarding its ability to process data. Shallow ML algorithms generally require the feature selection of relevant attributes for model training. This task can be non-trivial and time-consuming if the dataset is high-dimensional or if the context is not well-known to the model engineer. Common classes of shallow ML algorithms are linear regressions, decision trees, and support vector machines (SVM). Deep neural networks with multiple hidden layers and advanced neurons for automatic representation learning provide a computation- and data-intensive alternative (Janiesch et al., 2021; Mahesh, 2020). They master feature selection on increasingly complex data by themselves (LeCun et al., 2015; Schmidhuber, 2015). The performance of these deep-learning-based models surpasses shallow ML models and even exhibits super-human performance in specific applications such as data-driven maintenance (e.g., Wang et al., 2018) or medical image classification (e.g., McKinney et al., 2020). On the downside, the resulting models have a nested, non-linear structure, which is not per se interpretable for humans, and thus their predictions are difficult to retrace. In summary, many shallow ML algorithms are considered interpretable and, thus, white boxes, but deep learning algorithms tend to perform better but are non-transparent and, thus, black boxes (Adadi & Berrada, 2018; Wanner et al., 2022).

### 2.2. Interpretability and explainability in machine learning

In this context, interpretability signifies how accurately an ML model can associate cause and effect. It is an inherent, data-driven property that is related to the ML model's ability to provide meaning in understandable terms to a human by itself (Fürnkranz et al., 2020; Rudin, 2019).

In turn, explanations have the ability to fill the information gap between the intelligent system and its end user similar to the situation in the principal-agent problem (Arrieta et al., 2020; Baird & Maruping, 2021) whenever the ML model is non-transparent and therefore not sufficiently interpretable. Explanations are decisive for the efficacy of the intelligent system as the end user decides based on the given information whether he or she integrates the prediction into his or her decision-making or not (Shin, 2021; Thiebes et al., 2021).

The question of what constitutes explainability and how explanations should be presented to be of value to human users fuels an interdisciplinary research field, that consists of various disciplines, including philosophy, social science, psychology, computer science, and information systems (Collins et al., 2021; Miller, 2019). From a socio-technical perspective, explainability can be considered as the perceived quality of an explanation by an individual or user group (Adadi & Berrada, 2018; van der Waa et al., 2021). While the perceived quality can be circumstantial, we assume that there is a shared perception across user responses that can be used to explain at least part of the judgement.

From a technical point of view, explainability in intelligent systems is about two questions: the "how" question and the "why" question. The former is about global explainability, which provides answers to the ML algorithm's internal processing (Dam et al., 2018; Rudin, 2019). The latter is about local explainability, which answers the ex-post reasoning about a concrete prediction by an ML model (Arrieta et al., 2020; Dam et al., 2018). In this context, as noted above, many shallow ML models are considered to be white boxes that are interpretable per se (Arrieta et al., 2020; Janiesch et al., 2021). In contrast, a black-box ML model is either too complex for humans to understand or opaque for a reason and, therefore, equally hard to understand (Dwivedi et al., 2021; Rudin, 2019). Consequently, we consider an ML model's innate interpretability as its explainability towards end users not using any XAI augmentations (Adadi & Berrada, 2018; Kenny et al., 2021).

Theoretical contributions typically assume a continuous decrease in explainability with increasing performance of ML models. While it is generally depicted as a linear or cubic curve, it is not apparent whether this relation of explainability and performance is consistent across different ML models in a socio-technical evaluation with end users.

Likewise, there are numerous ways explanations can be presented with XAI. These augmentations can be summarized in six common explanation types (Mohseni et al., 2021).

*How* explanations represent a global view of the ML algorithm; common XAI augmentations display decision boundaries or model graphs. *Why* explanations represent a local view and describe why a prediction was made based on a singular input, demonstrating the importance of input variables for the decision of the model. Contrastive visualizations can be used to produce *Why-Not* explanations that outline the difference between an actual and the expected prediction. Furthermore, the algorithms' reaction to change in data or algorithmic hyperparameters can be outlined by *What-If* explanations. In a similar way, *How-To* or counterfactual explanations provide an interactive user experience, where the input of the model is changed in a way so that the output changes. Lastly, *What-Else* explanations offer explanations by example in providing training data that generate similar outputs from the model.

Aside from the type of explanation, it is essential to distinguish the target audience of the explanation. Research distinguishes four different groups: developers, theorists, ethicists, and users (Mohseni et al., 2021; Preece et al., 2018). As empirical findings differ for the stakeholder groups, we solely aim our study at the (end) user (Herm et al., 2021; Meske et al., 2022). In our case, end users are domain experts that use an intelligent system in their work routines to obtain predictions that assist their decision making. They do not participate directly in the system's planning, engineering, maintenance, or support and typically do not possess technical knowledge about its analytical model.

## 3. Theoretical background and hypotheses development

### 3.1. Machine learning tradeoffs

Considerations about the (hypothesized) tradeoff between model performance and model explainability have been the subject of discussion for some time. Originating from theoretical statistics, a distinction for different ML algorithms was first made between model interpretability and flexibility (James et al., 2013). More recently, this changed towards a comparison between model accuracy and interpretability (e. g., Arrieta et al., 2020; Yang & Bang, 2019) or algorithmic accuracy and explainability (e.g., Dam et al., 2018; Rudin, 2019). All tradeoffs address the same compromise of an algorithm's performance versus the algorithm's degree of result traceability.

Many subjective classifications of this tradeoff exist (Arrieta et al., 2020; Dam et al., 2018; Gunning, 2019; Nanayakkara et al., 2018; Rudin, 2019; Vempala & Russo, 2018; Yang & Bang, 2019). Overall, there is high conformity between the subjective classifications of the different authors. We synthesized these schemes into a generalized classification scheme.[5] The resulting Cartesian coordinate system shows five common classes of ML algorithms ordered by their assumed performance (*y*-axis) and explainability (*x*-axis).

There is a general agreement on key classes of ML algorithms, but there are some differences in their placement and the granularity of representation. The general notion is that with a loss of performance, algorithms provide better explainability so that algorithms can be ordered on some curve. Hence, deep neural networks are categorized as the most powerful algorithms with the least degree of explainability, followed by ensemble algorithms, which consist of multiple ML models. SVMs serve as a large margin classifier based on data point vectors and come third in performance, superior to decision trees that use sorted, aligned trees for the development of decision rules. Finally, linear regressions (or linear models in general) are considered least in performance yet straightforward to interpret (Goodfellow et al., 2016; James et al., 2013). Note certain classes of ML algorithms are thought to perform closer to each other than the conceptual equidistant visualization of Fig. 1 (e.g., Dam et al., 2018; Gunning, 2019; Guo et al., 2019).

In essence, these theoretical classification schemes represent a hypothetical and data-centered view on the tradeoff of model accuracy vs. model interpretability. They have neither yet been validated for specific applications based on real data nor confirmed by including end users to unearth their true pertinency towards said tradeoff between performance vs. explainability. An empirical quantification of end user explainability is necessary to provide first-hand knowledge to the engineers of intelligent systems for the development of intelligent systems (Jauernig et al., 2022; Meske & Bunde, 2022). Despite this apparent deficiency, they are commonly referenced as a motivation for a user- or organization-centered XAI research or intelligent system deployment (e. g., Asatiani et al., 2021; Guo et al., 2019; Rudin, 2019).

Augmented models (i.e., ex-post explainers) were the subject of several of those studies (e.g., Angelov & Soares, 2019; Nanayakkara et al., 2018). We have not included them in this synthesis as our focus was on the ML algorithms' inherent explainability. Yet, as noted above, XAI augmentations aim to provide more transparent ML models with both high performance and high explanatory power to improve the acceptance of predictions by end users (Arrieta et al., 2020; Gunning, 2019). Hence, it is self-evident that we need to consider how XAI augmentations can improve the explainability of the highest-performing – supposedly least explainable – ML algorithm.

There are human-centered evaluations of XAI algorithms. However, most of the evaluations revolve around testing certain novel algorithms. They often focus on variable importance and semi-automated evaluations by perturbating features to identify the most influential features and compare the result set with the XAI algorithms' explanations (Nguyen, 2018; Doshi-Velez & Kim, 2017). Thus, they measure the algorithms' explanatory performance towards a truth value rather than their explanatory quality towards end users.

In summary, it remains unclear how end users perceive explainability and how this is in line with the considerations presented above. We propose to approach this as follows: First, we focus on the tradeoff between performance and an ML model's inherent explainability to avoid biases introduced by model transfer techniques from the field of XAI. Second, we focus on five of Mohseni et al. (2021)'s six common types of explanation to augment the best-performing ML model to uncover which types of XAI explanations end users prefer independently of their potential to correctly explain an ML model or its predictions.

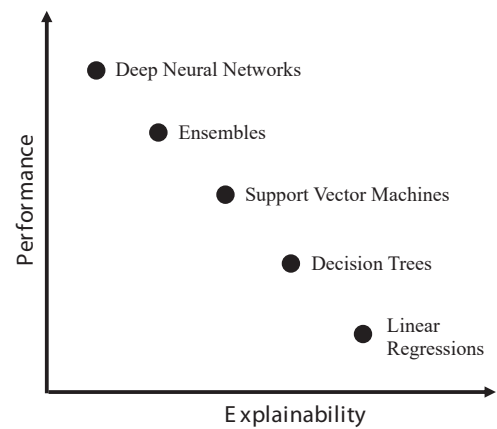

Fig. 1. Synthesis of Common ML Algorithm Classification Schemes.

---

[5] See Appendix A for an enlarged version of the figure, including the different classification schemes from literature.

### 3.2. Hypotheses

#### 3.2.1. Performance vs. explainability tradeoff

Our research design uses a simple group structure, where the independent variable is the *choice of algorithm,* and the dependent variables are *performance* and *perceived goodness of explanation*. The independent variable reflects the nature of an algorithm as it is applied to practical problems. The dependent variable performance measures the objective performance of the algorithm. The perceived goodness of explanation is more subjective and we base the choice of this second dependent variable on the proposed tradeoff that requires a quantification of explanation as it is perceived by users and knowingly can influence the user's mindset towards algorithms (Berger et al., 2021; Jauernig et al., 2022). The moderating group variable *data complexity* is expressed through different cases using different datasets reflecting *low complexity* and *high complexity*. We choose to introduce this variable to also reflect on more complex practical problems that involve large, non-tabular datasets like image and video data. These complex datasets are massive in size, high-dimensional, possibly biased, and not straightforward to explore by the human user. This also combats the predisposition that it is always viable to choose a simple algorithm that is explainable when clearly in those complex cases these types of algorithms are not interpretable as the data is neither (Castiglioni et al., 2021; Wang et al., 2021). When it comes to using algorithms for complex cases, post hoc XAI explanations can be used to provide insights into the decision-making process (Meske & Bunde, 2020; van der Waa et al., 2021) as they have an increased need for explainability (Lebovitz et al., 2021; Liu et al., 2007).

In terms of performance, we hypothesize that for less complex cases, which use tabular data, the performance of the ML models will be very close and not significantly distinguishable (Rudin & Radin, 2019; Zhang & Ling, 2018). Recent replication studies show that prediction scenarios using tabular data can be solved with small-scale models and will hold similar performance or even outperform the more advanced models for those low complexity cases (DeVries et al., 2018; Mignan & Broccardo, 2019). Thus, we expect no significant difference in performance following the ordering of Fig. 1.

**H1a**. *The choice of algorithm has no significant impact on the performance for cases with low complexity.*

Contrary, since shallow learning algorithms are limited in their way of extracting higher-level features for complex data, we expect the performance to deviate for complex cases (Janiesch et al., 2021; LeCun et al., 2015). Related research shows a decline in error rates for deep neural networks when applied to image datasets that even outperform human judgment (Heinrich et al., 2019; McKinney et al., 2020). Hence, we theorize that the performance of the shallow ML algorithms will be sub-par to deep neural networks and less grouped since they will fall off at different paces.

**H1b**. *The choice of algorithm has a significant impact on the performance for cases with high complexity.*

As a next step, we introduce the hypotheses regarding the goodness of explanations of common classes of ML algorithms. While we expect the black-box deep learning model to be the poorest in explainability (Meske et al., 2022; Rudin, 2019), we can only offer some thoughts on the ordering of shallow ML algorithms. First, we believe that the design of the explanation plays an important role in conveying the intended level of transparency (Miller, 2019; Shin, 2021). Shallow ML algorithm classes that have intrinsic means of local interpretability, such as SVM and linear regression in terms of their input feature weights, still have no natural way of visually presenting local variable importance out-of-the-box. The only exception are decision trees that present a logical structure, which is in line with the human thought process (Herm et al., 2021; Subramanian et al., 1992). Thus, we believe that contrary to the existing theories, only decision tree explainability will be distinguishable from the rest of the ML algorithms.

**H2a**. *The choice of algorithm has a significant influence on the perceived goodness of explanation for cases with low complexity.*

Following our argumentation, we believe that post hoc analysis will reveal that the significant overall difference can only be attributed to the difference between the decision tree and the other groups of algorithms for less complex cases.

Further, we believe this will not hold for the less complex case. Complex data structures like images, even when referred to by a tree structure, have no convincing explanatory value since there are so many input variables (in the image case: pixels) to choose from (Chandra & Bedi, 2021; Heinrich et al., 2019). For this type of data, an additional step is required to produce high-level features that humans can relate to, such as specific geometric forms.

**H2b**. *The choice of algorithm has no significant impact on the perceived goodness of explanation for cases with high complexity.*

#### 3.2.2. Perception of explanation types

For our second research question, we follow up on the first experiment to investigate the question of how the high-performing but black-box class of deep learning algorithms should be augmented with XAI explanation types. We aim to find out which type of explanation augmentation will help to elevate the transparency of deep learning models in the case of complex data. Thus, as independent treatment variables, we use the *type of explanation* (Mohseni et al., 2021). As a dependent variable, we again use the *perceived goodness of explanation*. Since deep learning algorithms come with no explanation at all, we theorize that any explanation will be favorable or equal to no explanation (Adadi & Berrada, 2018; Miller, 2019). Furthermore, we suspect that explanation types with straightforward, non-complex visualizations (e.g., *Why*) will be perceived more benevolently by end users. Thus, we formulate the following hypothesis:

**H3**. *The type of explanation has a significant impact on the perceived goodness of explanation.*

We believe that the distinction between more local-oriented (*Why*, *Why-Not*, and *What-Else*) and global-oriented types (*How* and *How-To*) will be notable. We grounded this expectation in our focus on end users. Developers and theorists need to understand the nature of the algorithm. End users are satisfied with simpler and more targeted example-based explanations of predictions that rationalize their belief in the system and assist in solving the task at hand (Miller, 2019; Preece et al., 2018). Fig. 2 summarizes the research models for RQ1 and RQ2. Note that both experiments stand alone, and that experiment 2 has a set choice of algorithm (deep neural network) and dataset (high complexity) and uses post hoc XAI augmentations rather than presenting the non-augmented, inherent explanations as in RQ1.

### 4. Methodology

#### 4.1. Performance vs. explainability tradeoff

To execute our research design for RQ1, we use a standard ML analysis process and subsequently conduct an empirical analysis (Müller et al., 2017). In our experiment, low data complexity is represented by a standard tabular dataset with a moderate number of observations and low dimensionality. High data complexity is represented by a large image dataset that exhibits high dimensionality and initial non-tabular form (i.e., pixel-tensors). See Table 1 for details on the datasets. Both datasets represent classification tasks.

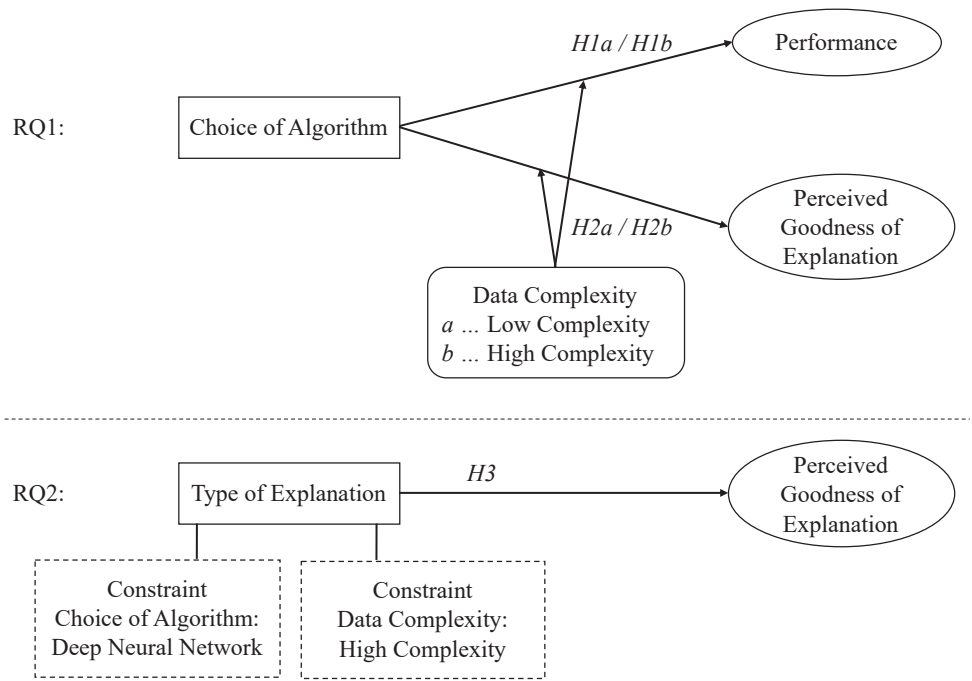

**Fig. 2.** Proposed Research Models for RQ1 and RQ2.

**Table 1**
Overview of Datasets and Moderating Effects.

| Dataset | Moderator Effect | Description |
|---|---|---|
| HEART (Janosi et al., 1988) | Low complexity | The heart disease dataset (HEART) is a low complexity dataset, which is used to classify the presence of heart disease based on medical patient features. It contains 303 observations of 13 different features and a binary target variable. |
| BRAIN (Bohaju, 2020) | High complexity | The brain tumor dataset (BRAIN) contains images of brain MRIs of which 2.079 depict no brain tumor and 1.683 depict a brain tumor. The images have dimensions of 224 × 224 pixels each. A binary label indicates the tumor status. |

For each dataset, we apply several classes of ML algorithms that represent the levels of the independent treatment variable *choice of algorithm* (see Table 2). We applied data preprocessing and grid search optimization for every ML algorithm. We ensured that each algorithm provides an acceptable answer for each case so that the class of algorithms can – in principle – be considered fully interchangeable from an end user perspective except for their performance and explanation.

**Table 2**
Overview of ML Algorithm Implementations.

| Class of ML Algorithm | Implementation |
|---|---|
| Linear Regression | Due to data preprocessing, we skipped default normalization and used the default settings. For the non-centered datasets, we included the intercept of the model. |
| Decision Tree | We did not restrict the models by regulations such as the minimum sample split numbers of the estimators. The resulting trees have a depth of five or six, depending on the treatment. |
| SVM | For all datasets, we applied an SVM using a radial basis function as kernels. |
| Random Forest (Ensemble) | We used the bagging algorithm random forest as a proxy for ensembles. Random forests consist of 100 estimators each, and their complexity was not restricted (see decision tree). |
| Deep Neural Network | For HEART, we used a multi-layer-perceptron with eight hidden layers, including dropout layers. For BRAIN, we used a convolutional neural network consisting of 13 hidden layers also including maxpooling and dense layers. |

After implementation and execution, we measure the dependent variable *performance* by measuring the accuracy of the classification by applying 15-fold cross-validation for each ML algorithm to ensure algorithm behavior in terms of reliability and possible variations. The accuracy of the model refers to the system's ability to correctly predict an outcome and is given by the ratio of correctly classified entities to all entities. The measure allows us to objectively estimate the performance without including perceived advice quality that can be biased by user perception (Janiesch et al., 2021; Mahmud et al., 2022). In addition, we reviewed recall and F-score to ensure that a single performance metric did not produce outliers.

While a model's performance can be evaluated independently of the user, its explainability depends on the perceptions of its users (Miller, 2019; Shin, 2021). Therefore, we measured the dependent variable, *perceived goodness of explanation,* in a survey. We used the platform prolific.co providing a monetary incentive. Since both our cases are healthcare cases and we aim our analysis at end users, we assume that the users of such systems would be healthcare professionals. Therefore, we used the filtering functionality of the platform to narrow down subjects to this group. Furthermore, to ensure a basic acceptance of AI among the group, we selected novice healthcare professionals (i.e., enrolled medical students) since experienced healthcare professionals can have a substantial bias to medical AI applications (Logg et al., 2019; Strohm et al., 2020). This also ensures the continuance of the results as those novices constitute the core of the future workforce. We find the notion of catering to as many groups as possible intriguing, but it is out of the scope of this research to consider technology acceptance as a factor (Straub & Burton-Jones, 2007).

For reasons of duration and repetitiveness, we designed two studies, one for each case. The cases were assigned at random. The procedure within each variant was identical. To ensure the validity and reliability of our study, we first asked a senior researcher for a review of our study. Then, we conducted a pre-study to check whether participants had understood the research design and the intended focus of the questions correctly. Furthermore, we asked about any difficulties encountered in completing the survey.

In the survey, we first collected demographics, prior experience with AI, as well as the participant's willingness to take risks. In the second part, we provided them with an introduction to the respective case and the task that the system is carrying out in that context. Third, we informed the participants that they had to put themselves in the situation of a physician who could not delegate the case. Then, we evaluated the participant's perceived goodness of explanation based on the propositions of Hoffman et al. (2018): We provided the participants with a graphical visualization of specific predictions dependent on the algorithm's inherent means to produce such explanations. Thereby, we account for each algorithm's natural way of explanation without adding further augmentations. For each ML model, the participants had to rate their overall perceived goodness of explanation of the model on a seven-point Likert scale. After yielding results for both dependent variables, to check on our hypotheses, we conducted an analysis of variance based on the design in Fig. 2.

To reduce participant bias, we applied different mechanisms and design elements. First, we randomized the order of treatments within every study to avoid any learning effects or sequence bias. Second, we did not use any colors or explanations from common ML implementation packages, as the participants could be biased through the presentation type (confirmation bias). Third, we only provide input information, an explanation, and a comprehensive description to each explanation, to not force anchoring biases. As an example, the participants did not receive any information about the performance of the ML models to avoid performance bias. Fourth, we assume no focus effects took place, as novice end users (generally) do not have actual prior experience in ML model explanations. Lastly, we applied a validation question as an awareness check. See Appendix B and C for the used study designs to answer RQ1.

### 4.2. Perception of explanation types

For RQ2, we adopted a similar approach to measure the quality of explanation. As a dataset, we used BRAIN as it represents the more complex dataset and applied only the deep neural network algorithm. Then, we implemented different XAI algorithms that reflect the levels of the independent treatment variable *types of explanation* levels as derived from Mohseni et al. (2021). We measured the *perceived goodness of explanation* analog to RQ1 by conducting a survey. We presented the case and the different explanations as a treatment to the participants. The treatment consists of three boxes: the original input image, the ML algorithm's decision (tumor/no tumor), a form of explanation, and a short textual description of the explanation to ensure a basic understanding. See Appendix D for the study design to answer RQ2.

Table 3 comprises the images presented as treatments to the participants for all explanation types. Please note that for our study, we combined Mohseni et al. (2021)'s explanation types *How-To* and *What-If* (in the following: *How-To*) as both focus on hypothetical adjustments to the input to generate what-if scenarios and counterfactual explanations. Due to our focus on end users, the wider scope of *What-If* explanations to include the changing model parameters is not applicable to our analysis. Furthermore, *What-If* explanations are not well suited for high-dimensional data and deep neural networks (Mohseni et al., 2021).

**Table 3**
Overview of Explanation Type Treatments for RQ2.

| Input Image | | |
|---|---|---|
| 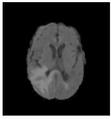 | | |

| Explanation Type | Description of Explanation Types | Treatment Visualization |
|---|---|---|
| *How* | Explanation of which input areas are relevant to the trained model, i.e., the ML algorithm's inner logic.[a] | 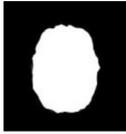<br>Result: Tumor |
| *Why* | Explanation of which areas of the given input are relevant to the outcome of the prediction.[a] | 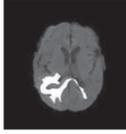<br>Result: Tumor |
| *Why-Not* | Explanation of which areas of the given input are not relevant to the outcome of the prediction.[a] | 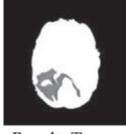<br>Result: Tumor |
| *How-To* | Explanation of how hypothetical adjustments of the given input (e.g., the bright shades in the MRI) would result in a (different) model prediction. | 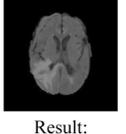 *Input Adjustment:* 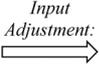<br>Result: Tumor    Result: No Tumor |
| *What-Else* | Explanation by example; showing similar inputs and their respective predictions. | 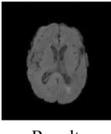 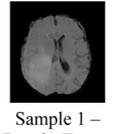 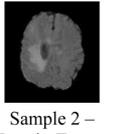<br>Sample 1 – Result: Tumor   Sample 2 – Result: Tumor   Sample 3 – Result: Tumor |

[a] The described XAI augmentations are highlighted as a white area in the treatment visualization.

## 5. Results

### 5.1. Result experiment i: performance vs. explainability tradeoff (RQ1)

#### 5.1.1. Participant demographics

In total, we received feedback from $n = 223$ subjects (HEART $n = 111$; BRAIN $n = 112$). To ensure the data quality of our findings, we excluded feedback by applying various preprocessing techniques, such as using control questions, detecting lazy patterns, deleting randomly filled questionnaires, and considering time constraints. This results in a final sample of $n = 100$ for HEART and $n = 101$ for BRAIN. Table 4 shows the demographics for both surveys.

#### 5.1.2. Performance

In general, the performance results support the theoretical ordering in Fig. 1 (*y*-axis). Nevertheless, the relative performance and thus the interval of the ordering differs. Especially, the difference between ensemble and SVM is more negligible than assumed. In our case, this may be due to the datasets and the ensemble algorithm. It reveals that the ordering of algorithms by their performance is as assumed in theory, but hardly deterministic.

Further, the performance difference between shallow ML algorithms and deep learning algorithms can be almost neglectable in scenarios with low complexity, such as HEART. Still, linear regression constantly performed worst while the deep neural network performed best. For the more complex case BRAIN, we encounter a strong decline in performance for all models except for the deep neural network. Table 5 illustrates the results of our performance evaluation derived through a mean calculation.

Using the folds from cross-validation, we executed a one-way analysis of variance (ANOVA) to check hypotheses H1a and H1b, respectively. Table 6 shows the results of the ANOVA. The full table with post hoc test results can be found in Appendix E, Table E.1.

Following the ANOVA, we cannot support H1a, but we can support H1b. The post hoc testing (see Appendix E, Table E.2) revealed the hypothesized differences between the two scenarios. In the case of low complexity, we found that linear regression significantly diverted from all other algorithms at the $p < 0.01$ level. We found no significant distinction for the other models.

**Table 4**
Descriptive Statistics of Subjects for Surveys from Experiment I (RQ1).

| Characteristics | Attributes | HEART[a] | | BRAIN[b] | |
|---|---|---|---|---|---|
| | | *Freq.* | *Percent.* | *Freq.* | *Percent.* |
| Gender | Male | 46 | 46.00 | 49 | 48.85 |
| | Female | 53 | 53.00 | 52 | 51.15 |
| | Others | 1 | 1.00 | - | - |
| Age (years) | < 20 | 13 | 13.00 | 20 | 19.80 |
| | 20–30 | 84 | 84.00 | 78 | 77.23 |
| | 31–40 | 2 | 2.00 | 3 | 2.97 |
| | 41–50 | 1 | 1.00 | - | - |
| Location | Africa | 19 | 19.00 | 19 | 18.81 |
| | Europe | 38 | 38.00 | 39 | 38.61 |
| | North America | 32 | 32.00 | 37 | 36.63 |
| | South America | 11 | 11.00 | 6 | 5.94 |
| Experience with AI (years) | None | 44 | 44.00 | 41 | 40.59 |
| | < 2 | 28 | 28.00 | 33 | 32.67 |
| | 2–5 | 19 | 19.00 | 19 | 18.81 |
| | 6–10 | 4 | 4.00 | 5 | 4.95 |
| | > 10 | 5 | 5.00 | 3 | 2.97 |

[a] $n = 100$
[b] $n = 101$

**Table 5**
Descriptive Statistics of *Performance* for *Choice of Algorithm*.

| Choice of Algorithm | HEART | | BRAIN | |
|---|---|---|---|---|
| | Mean Accuracy[a,b] | StdDev Accuracy[a] | Mean Accuracy[a,b] | StdDev Accuracy[a] |
| Linear Regression | 63.43 | 0.09 | 44.47 | 0.03 |
| Decision Tree | 73.86 | **0.06** | 57.98 | 0.03 |
| SVM | 82.34 | 0.09 | 65.66 | 0.08 |
| Random Forest (Ensemble) | 77.52 | 0.08 | 66.04 | **0.02** |
| Deep Neural Network | **84.42** | 0.11 | **89.45** | 0.04 |

[a] Calculations based on results from 15-fold cross validation
[b] higher = better, in %

**Table 6**
ANOVA Results for *Choice of Algorithm* and *Performance*.

| Dataset | Variable | Df[a] | Sum Sq[b] | Mean Sq[c] | F-value[d] | Pr (>F)[e] |
|---|---|---|---|---|---|---|
| Low complexity (HEART) | Choice of Algorithm | 4 | 0.4074 | 0.10186 | 16.85 | < 0.00001 |
| | Residuals | 70 | 0.4230 | 0.00604 | – | – |
| High complexity (BRAIN) | Choice of Algorithm | 4 | 1.4974 | 0.3743 | 165 | < 0.00001 |
| | Residuals | 65 | 0.1475 | 0.0023 | – | – |

[a] Degree of freedom
[b] sum squares
[c] mean squares
[d] result *F*-test
[e] result *p*-value

Observing the post hoc result for the complex case BRAIN yields another picture: While we can show the worst-performing role of linear regression in this scenario as well, the decision tree falls off as well and shows significant differences to all other models. Random forest and SVM exhibit nearly similar performance that does not distinguish significantly. In the top end, we find that deep neural network performance is a group of its own with significant distances to all other ML algorithms. In summary, we find that in the complex case BRAIN, the performance differences are more discernible and all ML algorithms except deep neural networks perform notably worse.

#### 5.1.3. Explainability

We present the perceived goodness of explanation from the conducted survey for each choice of algorithm in Table 7. We followed Boone and Boone (2012) and applied a median calculation for the Likert-type data. As the standard deviations appear normal with no natural anomalies, we applied an ANOVA for the results.

Across all treatments, random forests and decision trees achieved the highest or second-highest ratings. We show the results of the ANOVAs for perceived goodness of explanation in Table 8, and we can support H2a but we cannot not H2b.

For the low complexity case, we find the expected distribution of ML algorithms with interpretable models being superior in terms of explainability. Hence, the perceived explanation quality of the ML algorithms is significantly distinguishable with some notable exceptions: decision tree and random forest are perceived as similar, presumably due to both being based on tree algorithms and providing tree-structure visualization. In addition, we found that SVM and linear regression are perceived as equal when it comes to explanation goodness.

Surprisingly, for the complex case we find a similar picture. Although the perceived goodness of the decision tree has declined significantly, the perceived goodness between the groups SVM/linear regression and decision tree/random forest is still significantly distinguishable. The post hoc test indicates a strong deviation of the deep neural network from any other algorithm. The reason is straightforward as a deep neural network offers no inherent interpretability. It also shows that decision trees are still perceived as valuable explanations in complex cases.

**Table 7**
Descriptive Statistics of *Perceived Goodness of Explanation* for *Choice of Algorithm*.

| Choice of Algorithm | HEART | | BRAIN | |
|---|---|---|---|---|
| | Median Explainability[a] | StdDev Explainability[b] | Median Explainability[a] | StdDev Explainability[b] |
| Linear Regression | 4.00 | 1.31 | 3.00 | 1.14 |
| Decision Tree | **6.00** | 1.59 | **4.00** | 1.25 |
| SVM | 3.50 | **1.29** | 3.00 | 1.09 |
| Random Forest (Ensemble) | 5.00 | 1.59 | **4.00** | 1.21 |
| Deep Neural Network | 2.00 | 1.32 | 2.00 | **0.89** |

[a] Median of seven-point Likert scale: 1.00 = very low; 7.00 = very high
[b] standard deviation of seven-point Likert scale

**Table 8**
ANOVA results for *Choice of Algorithm* and *Perceived Goodness of Explanation*.

| Dataset | Variable | Df[a] | Sum Sq[b] | Mean Sq[c] | F-value[d] | Pr (>F)[e] |
|---|---|---|---|---|---|---|
| Low complexity (HEART) | Choice of Algorithm | 4 | 486.1 | 121.52 | 59.78 | < 0.00001 |
| | Residuals | 495 | 1006.3 | 2.03 | – | – |
| High complexity (BRAIN) | Choice of Algorithm | 4 | 343.7 | 85.92 | 68.28 | < 0.00001 |
| | Residuals | 500 | 629.1 | 1.26 | – | – |

[a] Degree of freedom
[b] sum squares
[c] mean squares
[d] result *F*-test
[e] result *p*-value

**Table 9**
Descriptive Statistics of Subjects for Survey from Experiment II (RQ2).

| Characteristics | Attributes | BRAIN-XAI[a] | |
|---|---|---|---|
| | | *Freq.* | *Percent.* |
| Gender | Male | 50 | 51.02 |
| | Female | 48 | 48.98 |
| Age (years) | < 20 | 17 | 17.35 |
| | 20–30 | 77 | 78.57 |
| | 31–40 | 4 | 4.08 |
| Location | Africa | 17 | 17.35 |
| | Europe | 41 | 41.84 |
| | North America | 35 | 35.71 |
| | South America | 5 | 5.20 |
| Experience with AI (years) | None | 40 | 40.82 |
| | < 2 | 33 | 33.67 |
| | 2–5 | 17 | 17.35 |
| | 6–10 | 4 | 4.08 |
| | > 10 | 3 | 3.06 |

[a] $n = 98$

### 5.2. Result experiment ii: perception of explanation types (RQ2)

#### 5.2.1. Demographics

For the second experiment, we obtained $n = 109$ responses using the high-complexity dataset BRAIN for the perception of different XAI explanation types (in the following, we refer to the sample as BRAIN-XAI) for the deep neural network. To ensure the data quality of our findings, we applied the same preprocessing techniques as in the first survey. The final sample consists of $n = 98$. The following table describes the demographics of the survey.

#### 5.2.2. Explainability

We present the perceived goodness of explanation from the survey for each type of explanation in Table 10. Since we used the same survey design as in the first experiment, we also applied a median calculation. Likewise, the standard deviations appear normal with no discernible anomalies.

Conducting the respective ANOVA, we can support H3 (see Table 11). Looking at the post hoc results (see Appendix F, Table F.1) in alignment with the descriptive statistics from Table 9, we find that there is a significant difference between no explanation and some sort of explanation, no matter what the type. The local explanations of *Why*, *What-Else*, *Why-Not* scored best, while the global explanation of *How* scored worst aside from the *baseline* of no explanation. Furthermore, we only find *Why-Not* and *How-To* explanations not significantly distinguishable.

## 6. Discussion and implications

### 6.1. Discussion

As the baseline for the discussion and the generalization of our findings to analyze the tradeoff between performance and explainability (RQ1), we have merged all data from the first experiment. We normalized the data to the range of 0–1 to allow for a relative comparison of the ML algorithms regarding the different use cases. We transferred our findings into a Cartesian coordinate system as in Fig. 1 to visualize our results next to the theoretical assumption. We used mean values to yield a position for each algorithm. Fig. 3 shows the resulting averaged scheme calculated from the data in Tables 5 and 7. It mirrors the results of (Wanner et al., 2021a).

We can support some tendencies mostly concerning ML model performance as reflected by accuracy. A few things are notably different from the theoretical proposition. In particular, the hypothetical curve between ML model performance and ML model explainability assumed by prior research (left) does not hold in our user-centered treatments (right).

As a result, our empirical evidence shows a grouped structure and challenges the assumption of a tradeoff curve. It is visible in both treatments in terms of explainability. We find that the tree-based models, such as decision trees and random forests, are perceived to provide the best explainability of the five ML models from an end user's perspective. While random forests fall into the ensemble class, the base class model for the ensemble is the decision tree. This explains the random forest's comparably high scores despite being an ensemble algorithm. Contrary to our expectations, we could not substantiate that a single decision tree is perceived as substantially more explainable than a random forest with many unbalanced decision trees. We assume that this may be the case since we did not present all resulting trees of the random forest to the participants for review.

**Table 10**
Descriptive Statistics for *Perceived Goodness of Explanation* for *Type of Explanation*.

| Explanation Type[a] | Median Explainability[b] | StdDev Explainability[c] |
|---|---|---|
| Baseline (Black Box) | 2.00 | 1.13 |
| How | 3.00 | 1.94 |
| Why | **6.00** | 1.43 |
| Why-Not | 5.00 | 1.87 |
| How-To | 4.00 | 1.19 |
| What-Else | **6.00** | 1.35 |

[a] All augmentations are based on same CNN model (acc: 89.45%)
[b] median of seven-point Likert scale: 1.00 = very low; 7.00 = very high
[c] standard deviation of seven-point Likert scale

**Table 11**
ANOVA results for *Type of Explanation* and *Perceived Goodness of Explanation*.

| Dataset | Variable | Df[a] | Sum Sq[b] | Mean Sq[c] | F-value[d] | Pr (>F)[e] |
|---|---|---|---|---|---|---|
| High complexity (BRAIN-XAI) | Explanation Type | 5 | 960.5 | 192.11 | 82.94 | < 0.00001 |
|  | Residuals | 582 | 1348.0 | 2.32 | – | – |

[a] Degree of freedom
[b] sum squares
[c] mean squares
[d] result *F*-test
[e] result *p*-value

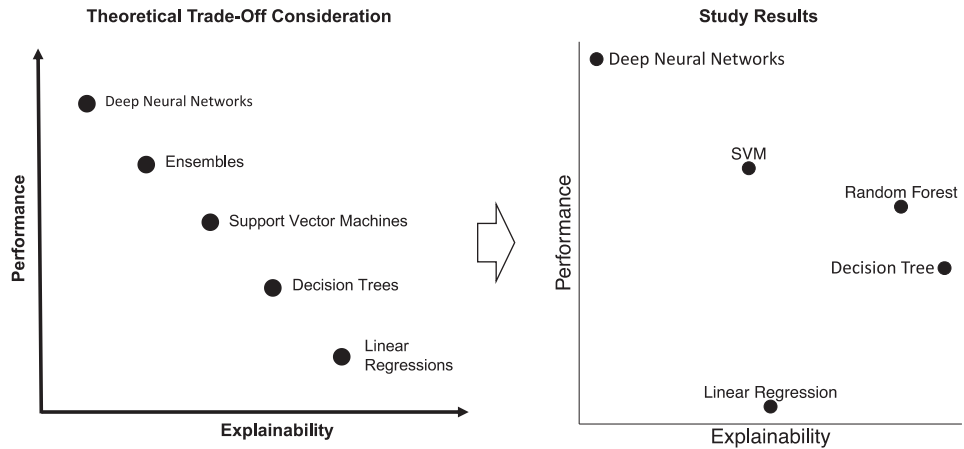

**Fig. 3.** : Theoretical vs. Empirical Scheme for the Tradeoff of Performance vs. Perceived Explainability in Machine Learning.

This observation provides new knowledge about the perceived explainability of ML algorithms and renders a more realistic picture of the performance vs. explainability tradeoff than the predominantly theoretical discussion considered the state-of-the-art (e.g., Arrieta et al., 2020; Gunning, 2019; Rudin, 2019).

Fig. 4 shows the non-normalized and normalized results for the two cases.

Continuing with the non-normalized schemes, we can also see that the shallow ML algorithms are positioned relatively close together in terms of their performance. For low complexity datasets, performance gains of deep learning over shallow ML are neglectable. Performance only becomes a factor for high complexity datasets. Hence, as expected, the performance distance between all algorithms widens with increasing dataset complexity and, thus, the choice of algorithm has a larger impact on an intelligent system's performance. Notably, we can also see that the absolute perceived explainability of tree-based algorithms wanes in the non-normalized schema with increasing dataset complexity, and the overall explainability distance decreases.

Consolidating both axes, we find the tradeoff to be dependent on the complexity of the case as well as on real-world performance and explainability requirements. As performance behaves differently than explainability, the tradeoff is non-trivial and consequently a multi-criteria decision (Gunning, 2019; Meske et al., 2022; Wanner et al., 2020). We provide further evidence of this cause.

Since XAI augmentations can be used to provide post hoc explanations of predictions, the tradeoff becomes even more complex as it becomes evident that – at least for certain applications – the use of high-performing deep learning algorithms may become an option despite their lack of explainability. To investigate how one can best augment these algorithms for end users, we implemented five common types of explanation in a subsequent survey (RQ2).

In Fig. 5, we summarize the perceived goodness of explanation regarding the explanation types on the employed Likert scale.

We can clearly see the local explanations such as *Why* and *Why-Not* are superior in comparison to the global explanation type *How*. Pointing out the tumor in a direct manner seems to resonate most with the end user as it supports their perception of the tumor's size and location. Even though similar, *Why-Not* explanations that require higher cognitive effort received lower scores than *Why* explanations. A similar argument can be made for *How-to* explanations as those explanations reference a decision to certain input changes to convey an understanding of the decision behavior. Surprisingly, the design of the *How-To* explanation seems to be perceived more benevolently than the *How* explanation. Lastly, *What-Else* explanations are well-received and show that it is beneficial to provide examples for the user to check whether the decision behavior of the algorithm is in line with his or her personal expectancy.

Across all types, we found that end users prefer local explanations that explain the result of a prediction either by pointing to a reason (*Why*) or giving examples (*What-Else*). Inversely, end users tend to disdain global explanations that merely visualize where the algorithms look (*How*) or show the mechanism behind a prediction (*How-To*). The results for contrastive explanations (*Why-Not*) were ambiguous, indicating that there may be applications for this kind of explanation, but while end users prefer them over global explanation, they generally favor non-negated reasoning.

With our socio-technical analysis, we provide first-hand evidence that the design of the explanation (i.e., how it is presented) plays an important role besides the content that is displayed (Mualla et al., 2022; Das & Rad, 2020). We also provide indicative knowledge of which types of explanation are more suitable for end users.

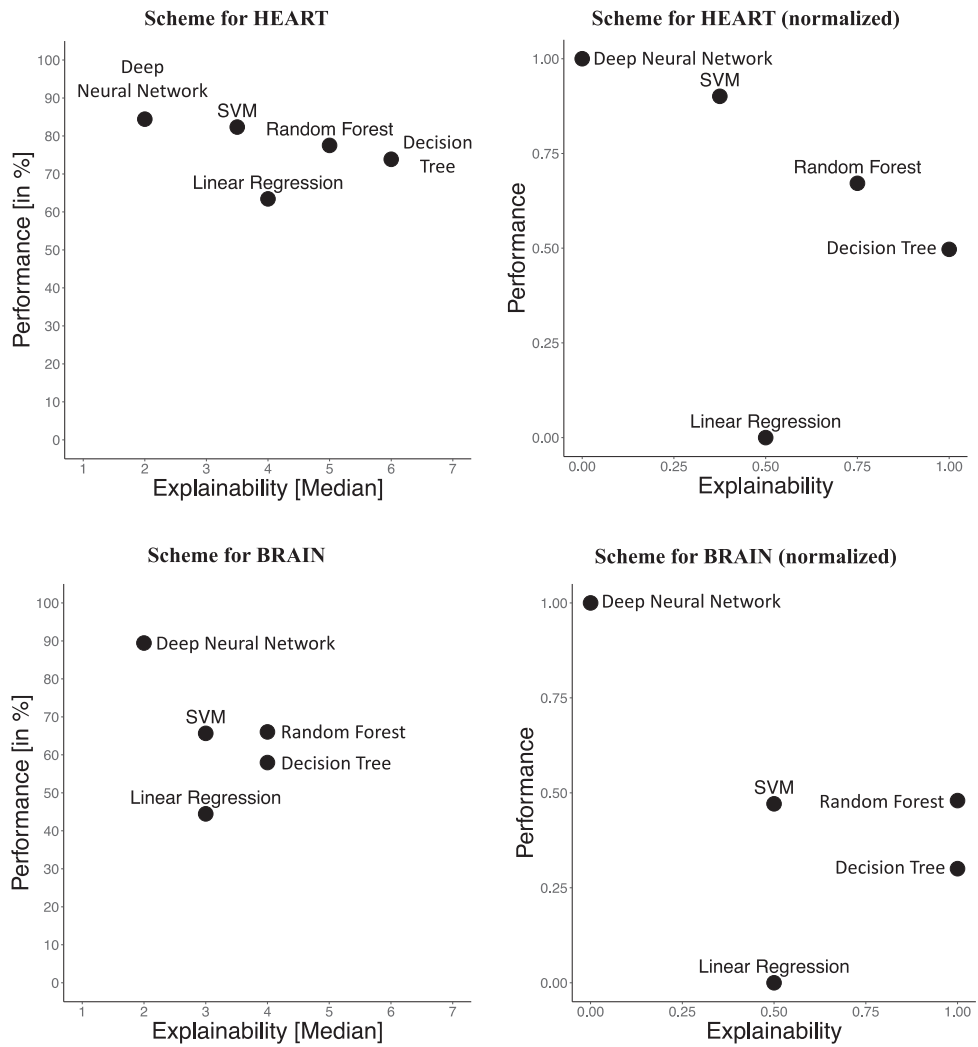

**Fig. 4.** Non-normalized and Normalized Empirical Schemes for the Tradeoff Between Performance and Explainability for Both Cases (HEART and BRAIN).

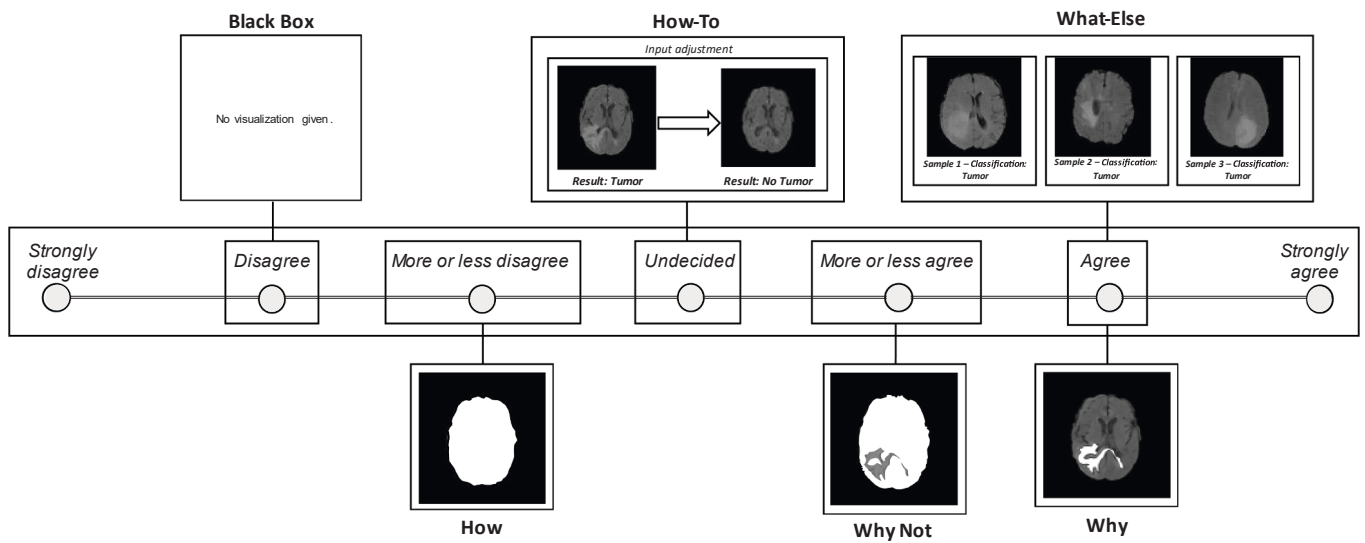

**Fig. 5.** Average End User Ratings of *Explanation Type* Visualized on Likert Scale.

## 6.2. Theoretical Implications

Our in-depth discussion focused on three major observations: the generalizability of results across the treatments, the relation between assumed model interpretability and perceived explainability, and end user preferences for explanation types. This allows us to summarize the key points of our theoretical findings as follows:

*6.2.1. Tackling the tradeoff between performance and explainability is non-trivial*

We showed that the tradeoff curve assumption between performance and explainability does not always hold. While we cannot prove that the relationship always exhibits a grouped structure, other evidence points to the fact that the tradeoff can be characterized as a group decision-making process where explanation and performance cannot be approached in isolation but in alignment with organizational policy and external factors such as laws (Ebers, 2020; Goodman & Flaxman, 2017). The decision process is also influenced by the perception of the system by the decision maker. Especially in intelligent systems that use ML algorithms as a basis for decision making, a plethora of individual factors like self-efficacy, general distrust, or neuroticism can influence the view of the tradeoff (Mahmud et al., 2022; Zhou et al., 2021). It requires a weighted multi-criteria decision process to represent and quantify elements for both dimensions, performance, and explainability (Meske et al., 2022; Wanner et al., 2020), which is complicated by increasing and decreasing distances between the respective algorithms. Furthermore, identifying decision elements is a challenge as some effects may only be detectable as latent indirect factors (Wanner et al., 2021b). Lastly, it is important to note if enhanced explainability does not translate into firm productivity, investing in XAI may be in vain for businesses. Further research using mixed-method approaches, including qualitative studies, can provide more detailed insights into end users valuations and avoid an explainability paradox.

*6.2.2. Model-inherent interpretability does not entail explainability*

The discrepancy between what is assumed in theory and our empirical findings can be explained at least in parts by the nature of our observations. Theoretical contributions look at the algorithmic and mathematical description of objects (data-centered perspective). We have employed a socio-technical and thus user-centered perspective. In our study, we targeted the naturally biased perception of end users of an ML algorithm directly and found that the difference between performance and explainability is not constantly increasing. Instead, we found that there are three groups of perceived explainability: none (deep neural networks), mediocre (coefficient-based algorithms such as linear regressions and SVM), and high (tree-based algorithms such as decision trees and random forests). The former group represents the concept of deep learning. The latter two represent shallow ML.

Decision trees are considered highly interpretable by humans in terms of their global and local explainability since it is possible to retrace a path of variables from the root node to a leaf node containing the final decision (Arrieta et al., 2020; Herm et al., 2021). This explainability by design makes the model itself (global) as well as every prediction (local) intuitively accessible. The similarity of coefficient-based models can be explained by the fact that both offer variable weights in the form of coefficients and, hence, any visualization that can be done for SVM would be valid for visualizing linear regression coefficients as well. This suggests that the goodness of the explanation can be attributed largely to its design and, thus, the basic type of the explanation rather than the actual content of the explanation. Lastly, deep neural networks are considered to be black boxes to the end users that are not interpretable by humans. They need to be augmented with XAI to offer any explainability to end users.

*6.2.3. Explanations and XAI augmentations that require low cognitive effort fare better*

End users clearly indicated that they perceive local explanations as more explainable. From this, we infer that people prefer explanations that require less cognitive effort to process and translate into their mental model. Tree-based algorithms that offer intuitive text-based access are perceived as more explainable than the competition. They combine local post hoc explanations with global decision process knowledge. This is in line with the observations of algorithm aversion theory, where it is required to see how the algorithm behaves to form a proper judgment (Berger et al., 2021; Jussupow et al., 2020). In line with Miller (2019), we also observed that end users prefer straightforward XAI augmentations, which require low cognitive effort such as local *Why* and *Why-Not* explanations, over complex global explanations such as *How* or *How-To*. Explanations that show decision examples, such as *What-Else* explanations, have a positive influence on the perception of explanations. It is important to note that the experiment only showed reactions to initial exposure and did not consider learning that occurs through either teaching or experience. The learning effects might reduce cognitive effort and make other means of explanation more accessible.

*6.3. Practical implications*

Our analysis and theoretical findings bear relevance for business practice as we can derive several practical implications from observing how people perceive algorithmic advice and explanations. Below, we summarize our findings by providing three guidelines that should be considered when employing an intelligent system for a specific task:

*6.3.1. Start with the performance threshold*

If an analytical model does deliver the required performance, it is not fit for the task. An explainable model that cannot provide the requested minimum quality will have no value in practice. Hence, all candidate algorithms must fulfill this requirement. The threshold will usually be determined by the overarching goal of the system involving business goals (e.g., cost savings) that can be realized by using the system.

*6.3.2. Consider organizational or project context beyond performance*

Other constraints typically influence the choice of algorithm. They largely depend on environmental factors such as cost (training and inference), time constraints, end user abilities, and laws. For example, models with high inference times cannot be used in real-time settings (e. g., defect detection in production). Tree-based algorithms are particularly accessible for ML laypeople, while coefficient-based algorithms may provide better performance and still be explainable enough for a trained workforce. Furthermore, laws such as the GDPR could require you to implement either a per se interpretable algorithm or a post hoc XAI augmentation. Therefore, the candidate pool must deliver acceptable explanations not only to end users but potentially also to the authorities. In that regard, consider Rudin's call for using inherently interpretable models whenever possible and keep note that the performance gain through deep learning can be neglectable for low complexity datasets (Rudin, 2019).

*6.3.3. Consider the degree of explanation that end users need*

Do not confuse model interpretability (required by experts to analyze the decision-making process) and prediction explainability (necessary for end users to make decisions in their work processes). After deploying an intelligent system, end users will use the system to fulfill their work tasks and not to analyze the model's decision process. These end users should be included in the explanation design process (explanation type, but also colors, visuals, etc.). This ensures that the explanations are appropriate for end users to assess the quality of the system's predictions and consider its advice appropriately. For novice users of the system, local *Why* and *What-Else* explanations promise the best user acceptance. In contrast, more global *How* and *How-To* explanations require more cognitive effort but may help them to better understand the decision process and gather expertise faster.

*6.4. Limitations and future research directions*

As with any empirical research, our study faces some limitations.

First, our study uses online surveys with benchmarking datasets. While we only allowed for participants with a certain background,

participants may have been exposed to the scenarios and several of the ML algorithms for the first time. Hence, we measured an *initial* explainability. Both datasets stem from the healthcare sector. This may introduce bias. In response, we have piloted similar surveys in different domains with comparable results (Wanner et al., 2021a). Furthermore, there was no time constraint for viewing and assessing an explanation. We expect results to differ in a high-velocity treatment where faster inference time becomes more valuable. Moreover, we compared inherently interpretable shallow ML algorithms and deep neural networks without further augmentations. We assume that XAI augmentations will affect explainability positively and initial evidence points to the fact (Herm et al., 2021), but we refrained from including it in the first experiment due to the diversity of explanation types and visualization options that we only began to explore in the second experiment. Nevertheless, a comprehensive evaluation of the explainability of XAI augmentations is necessary to gain a better understanding. This would include assessing whether single, isolated explanations work best or if users should be presented with explanations in pairs, triplets, or even more explanation types at the same time.

Second, choosing an ensemble model (in our case, random forest) always yields bias toward the interpretability of the base algorithm class of the ensemble model. In our case, the choice of random forest caused an overestimation of ensemble explainability due to the high degree of explainability of decision trees. Consequently, we expect other ensemble models to perform consistently according to their respective algorithm base classes. To evaluate this, we suggest testing multiple ensemble models that use a variety of base class permutations to give a more objective overview of ensemble explainability. Performing such an analysis was out of scope for our research.

Third, it is possible that participants were biased in their judgment by the perceived capability or promise of algorithms and therefore assumed a higher value (Hilton, 1996; Mehrabi et al., 2021). That is, shallow ML algorithms such as SVM and linear regression offer a form of internal explainability. Hence, they were supposed to result in a better-perceived explainability than black-box models with no internal explainability, such as deep neural networks. However, we found that difference to be smaller than expected. This may be due to participants who were not able to understand the presentation of SVM and linear regression as they lacked prior knowledge (Amershi et al., 2019; Arrieta et al., 2020), which may be a practical problem in real-life cases as well. Due to high expectancy in one category (performance), end users may attribute higher scores in another category (explainability), resulting in a halo effect. Furthermore, the perceived overall impression of an algorithm can be attributed to other factors that were omitted from the study in a controlled manner. Lastly, we did not measure whether the use of certain ML algorithms or XAI augmentations improved end user task performance and thus productivity. We assume a correlation between perception, understanding, and task performance as Herm et al. (2021) report. However, we did not directly measure this.

## 7. Conclusion

Despite its fundamental importance for human decision-makers, empirical evidence regarding the tradeoff between ML model performance and explainability is scarce. In response, we conducted an empirical study to develop a more realistic understanding of this relationship (RQ1) and explore the effect of various explanation types on end users (RQ2).

We underscore that existing theoretical propositions on the tradeoff are data-centered and misleading oversimplifications in terms of end user explainability. You cannot exchange performance for explainability and vice versa in a continuous fashion. Rather than a tradeoff curve assumption, we found a grouped structure of no, mediocre, and high explainability, where the explanation quality of decision trees and random forests constantly dominates other ML models. Further, we found that explanations fare better when they require less cognitive effort such as local explanations.

In our research, we measured the naturally biased perception of explanations by end users and not their understanding, learning effects, or task performance. Research into the usefulness of AI and human biases in ML is still in its infancy and requires substantial advances to pinpoint the effects of the various factors in play.

## CRediT authorship contribution statement

**Lukas-Valentin Herm:** Conceptualization, Methodology, Validation, Formal analysis, Investigation, Data curation, Writing – original draft, Writing – review & editing, Visualization. **Kai Heinrich:** Conceptualization, Methodology, Validation, Formal analysis, Investigation, Data curation, Writing – original draft, Writing – review & editing, Visualization. **Jonas Wanner:** Conceptualization, Methodology, Validation, Formal analysis, Investigation, Data curation, Writing – original draft. **Christian Janiesch:** Conceptualization, Methodology, Resources, Supervision, Writing – original draft, Writing – review & editing.

## Acknowledgement

This research and development project is funded by the Bayerische Staatsministerium für Wirtschaft, Landesentwicklung und Energie (StMWi) within the framework concept "Informations- und Kommunikationstechnik" (grant no. DIK0143/02) and managed by the project management agency VDI+VDE Innovation + Technik GmbH.

## Appendix A. Supporting information

Supplementary data associated with this article can be found in the online version at doi:10.1016/j.ijinfomgt.2022.102538.